\documentclass[sigconf]{acmart}
\usepackage{hyperref}
\usepackage{soul}
\usepackage{ulem}
\usepackage[anythingbreaks]{breakurl}
\usepackage{multirow} 
\usepackage{fancyhdr}
\usepackage{subcaption}
\usepackage{booktabs, tabularx, adjustbox, ltablex}
\usepackage{amsmath}
\newtheorem{Definition}{Definition}
\newtheorem{Theorem}{Theorem}
 
\newtheorem{definition}{Definition}[section]
\usepackage{float}
\usepackage{algorithm}
\usepackage{algpseudocode}
\acmDOI{https://doi.org/10.1145/3711896.3736904}
\acmISBN{}
\acmConference[KDD`25]{Make sure to enter the correct
 conference title from your rights confirmation email}{August 03-07, 
 2025}{Toronto, Canada}
\setcopyright{acmcopyright}
\acmSubmissionID{ }
 
\begin{document}
%
% The "title" command has an optional parameter, allowing the author to define a "short title" to be used in page headers.
% \title[STH-SepNet-GNN]{STH-SepNet-GNN: Leveraging Hypergraph Learning for Precise and Scalable Traffic Demand Forecasting with Large Language Models}
\title[STH-SepNet]{Decoupling Spatio-Temporal Prediction: When Lightweight Large Models Meet Adaptive Hypergraphs}

%
% The "author" command and its associated commands are used to define the authors and their affiliations.
% Of note is the shared affiliation of the first two authors, and the "authornote" and "authornotemark" commands
% used to denote shared contribution to the research.
% Authors
\author{Jiawen Chen}\authornotemark[1]
\email{jiawenchen@seu.edu.cn}
\affiliation{%
  \institution{Southeast University}
  \city{Nanjing}
  \country{China}
}
 
\author{Qi Shao}
\email{mathshaoqi@seu.edu.cn}
\affiliation{%
  \institution{Southeast University}
  \city{Nanjing}
  \country{China}
}

\authornote{Jiawen Chen and Qi Shao contributed equally to this work.}
 
\author{Duxin Chen}\authornotemark[2]
\email{chendx@seu.edu.cn}
\orcid{}
\affiliation{%
  \institution{Southeast University}
  \city{Nanjing}
  \country{China}
}
 
\author{Wenwu Yu}
\email{wwyu@seu.edu.cn}
\affiliation{%
  \institution{Southeast University}
  \city{Nanjing}
  \country{China}
}

\authornote{Correspondence to Duxin Chen and Wenwu Yu.}

% }
% \renewcommand{\shortauthors}{Toronto, Canada}

%
% The abstract is a short summary of the work to be presented in the article.
\begin{abstract}
Spatio-temporal prediction is a pivotal task with broad applications in traffic management, climate monitoring, energy scheduling, etc. However, existing methodologies often struggle to balance model expressiveness and computational efficiency, especially when scaling to large real-world datasets. To tackle these challenges, we propose STH-SepNet (Spatio-Temporal Hypergraph Separation Networks), a novel framework that decouples temporal and spatial modeling to enhance both efficiency and precision. Therein, the temporal dimension is modeled using lightweight large language models, which effectively capture low-rank temporal dynamics. Concurrently, the spatial dimension is addressed through an adaptive hypergraph neural network, which dynamically constructs hyperedges to model intricate, higher-order interactions. A carefully designed gating mechanism is integrated to seamlessly fuse temporal and spatial representations. By leveraging the fundamental principles of low-rank temporal dynamics and spatial interactions, STH-SepNet offers a pragmatic and scalable solution for spatio-temporal prediction in real-world applications. Extensive experiments on large-scale real-world datasets across multiple benchmarks demonstrate the effectiveness of STH-SepNet in boosting predictive performance while maintaining computational efficiency. This work may provide a promising lightweight framework for spatio-temporal prediction, aiming to reduce computational demands and while enhancing predictive performance. Our code is avaliable at  
https://github.com/SEU-WENJIA/ST-SepNet-Lightweight-LLMs-Meet-Adaptive-Hypergraphs. 
\end{abstract}


% This command processes the author and affiliation and title information and builds
% the first part of the formatted document.
\maketitle

%-------------------------------------------------------------------------------

\section{Introduction}\label{}
 
Spatio temporal prediction serves as a fundamental component of modern data-driven decision-making, with applications in urban traffic forecasting~\cite{wu2020connecting, choi2022graph}, climate modeling~\cite{nguyen2023climax, bi2023accurate}, and energy grid optimization, etc. Despite its broad significance, the field faces two primary challenges: accurately capturing dynamic spatial dependencies and ensuring computational scalability for large-scale real spatio-temporal datasets~\cite{yan2023spatio}. While deep learning has led to notable advancements, existing methods often struggle to achieve a balance between model expressiveness and computational efficiency~\cite{OpenCity}.

Recent advances in graph neural networks (GNNs) and large language models (LLMs) have been extensively explored to address these challenges~\cite{kipf2016semi, touvron2023llama}. GNNs are particularly effective in capturing spatial dependencies through graph-structured representations~\cite{yuan2024unist}. 
However, their dependence on static graph topologies poses a significant constraint, impeding their capacity to accurately model dynamic, higher-order interactions. For instance, in traffic networks, the influence between regions is constantly evolving and influenced by external conditions—factors that are inadequately captured by static adjacency matrices. Meanwhile, LLMs, which distinguish themselves in temporal prediction due to their strong sequence modeling capabilities~\cite{li2024gpt}, incur substantial computational expenses when applied to large node sets. Furthermore, their ability to leverage spatial structures remains limited. The dominant approach of jointly modeling spatial and temporal features within a single framework has been shown to exacerbate these challenges, often leading to overparameterized models that are computationally demanding and difficult to optimize, without yielding proportional performance improvements~\cite{jin2023time}. This raises a critical question: Can spatial and temporal modeling be decoupled to achieve both efficiency and accuracy?

\begin{figure}[t]
  \centering
  \begin{subfigure}[]{\linewidth}
    \centering
    \includegraphics[width=0.85\linewidth]{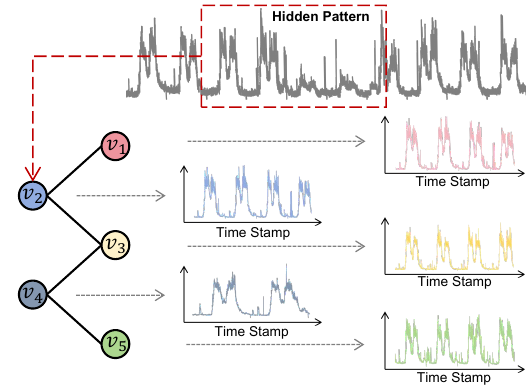}
    % \caption{Spatial Drift of Data}
    \label{fig:fig1a}
    \vspace{-10pt}
  \end{subfigure}
  \begin{subfigure}[]{\linewidth}
    \centering
    \includegraphics[width=0.85\linewidth]{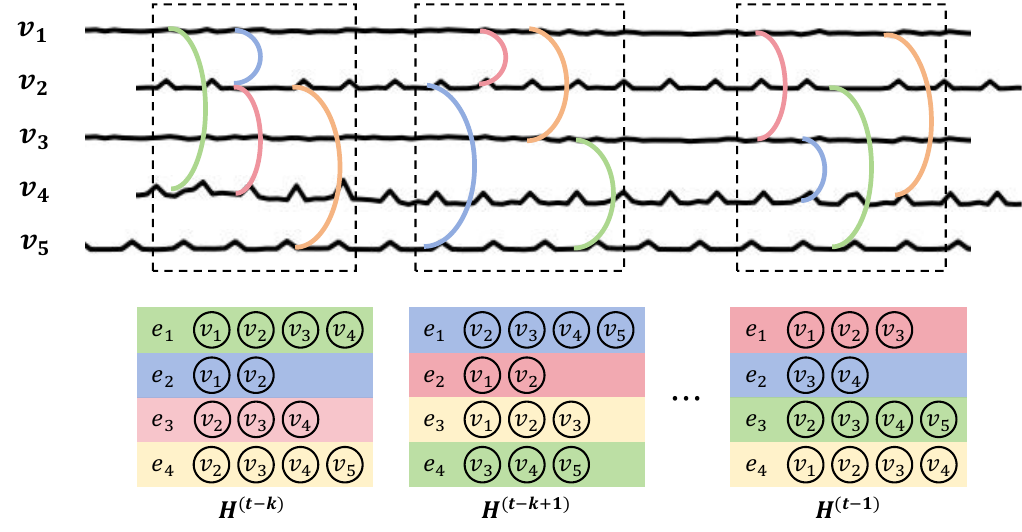}
    % \caption{Dynamic Adaptive Hypergraph}
    \label{fig:fig1b}
    \vspace{-10pt}
  \end{subfigure}
  \caption{(a) Spatio-temporal data exhibit spatial distribution shifts across different nodes. (b) Dynamic adaptive hypergraph captures evolving spatial distribution patterns.}
  \label{fig01}
  \vspace{-18pt}
\end{figure}

To tackle these challenges, this work proposes a separation strategy based on two key insights. First, as illustrated in Figure~\ref{fig01}, temporal dynamics in spatiotemporal systems often exhibit a low-rank structure, implying that the evolution of system states can be efficiently characterized by a small number of latent factors~\cite{bahadori2014fast, nie2024imputeformer}. This low-rank property facilitates the use of lightweight sequence models, such as distilled versions of LLMs, to capture temporal trends without compromising expressiveness. Second, spatial dependencies in spatiotemporal systems can be viewed as a form of spatial drift, where the influence between nodes shifts over time due to external factors or intrinsic system dynamics. Traditional GNNs, despite their effectiveness in many applications, struggle to capture dynamic drift due to their dependence on static graph structures~\cite{wu2020connecting, song2020spatial}. To address this limitation, we propose an adaptive hypergraph framework that possesses enhanced representational capabilities for graphs, enabling it to model evolving higher-order interactions. This framework allows hyperedges to dynamically encapsulate shifting relationships among multiple nodes, thereby accurately reflecting the evolving nature of these connections.

Building on these insights, we propose STH-SepNet (Spatio-Temporal Hypergraph Separation Network), a novel framework that specifically integrates lightweight temporal modeling with adaptive spatial modeling. For temporal modeling, compact LLMs (e.g., BERT~\cite{devlin2018bert}, GPT-2~\cite{radford2019language}) are employed to efficiently capture low-rank temporal dynamics. For spatial modeling, an adaptive hypergraph neural network is introduced to dynamically construct hyperedges, enabling the representation of spatial drift and higher-order interactions. A gating mechanism is further designed to fuse temporal and spatial representations, ensuring seamless integration while maintaining computational efficiency. The main contributions of this work are summarized as follows:
\begin{itemize}
\item We propose a lightweight spatio-temporal separation framework (STH-SepNet) for spatio-temporal prediction tasks. The framework integrates textual information and latent spatial dependencies, resulting in significant improvements in predictive performance.
\item We design an adaptive hypergraph structure for spatial modeling, which dynamically constructs complex dependency relationships and enhances the extraction of spatial features through effective order modeling.
\item We conduct extensive experiments to validate STH-SepNet, demonstrating state-of-the-art performance across multiple benchmarks. The proposed method demonstrates efficient execution on a single A6000 GPU, underscoring its practical applicability for real-world deployment.
\end{itemize}

\section{Related works}
%\textcolor{red}{DONE} 
\subsection{Large Models for Prediction} 
Large language models, recognized for their extensive parameter sizes and strong generalization capabilities~\cite{goodge2025spatio}, have been increasingly applied to time-series analysis tasks, including prediction, classification, and imputation. To bridge the gap between numerical data and the text-based processing paradigm of LLMs, researchers have explored novel data formatting techniques. For instance, PromptCast converts numerical sequences into natural language prompts~\cite{xue2023promptcast}, while Gruver et al. encode time-series data as digit strings to enable zero-shot predictions~\cite{gruver2024large}. These approaches demonstrate the potential of LLMs in temporal modeling while also highlighting the need for specialized adaptations to address the unique challenges of time-series data, such as irregular time intervals and long-range dependencies.

Recent efforts have sought to refine tokenization and embedding strategies to improve LLMs' suitability for forecasting tasks. LLM4TS employs parameter-efficient fine-tuning (PEFT) to adapt pre-trained LLMs for time-series prediction~\cite{chang2023llm4ts}, while Zhou et al. propose a unified framework for handling diverse time-series tasks~\cite{zhou2023one}. Additionally, advancements such as reprogramming frameworks~\cite{jin2023time} and contrastive embedding strategies~\cite{li2024unicl} further align numerical and textual modalities, enhancing LLMs' capability to process temporal data. However, these methods predominantly focus on temporal modeling while largely overlooking spatial dependencies, which poses a fundamental limitation for spatio-temporal prediction tasks.

The integration of LLMs with transformer-based architectures has further expanded their applicability to domain-specific challenges. Models such as UniST~\cite{yuan2024unist} and OpenCity~\cite{OpenCity} incorporate transformers with graph neural networks to capture spatio-temporal dependencies, while ClimaX~\cite{nguyen2023climax} and Pangu-Weather~\cite{bi2023accurate} illustrate the versatility of transformer-based designs in climate forecasting. Despite these advancements, existing approaches often struggle to effectively balance spatial and temporal modeling, resulting in increased computational complexity without proportional performance improvements. This underscores the need for novel frameworks that integrate spatio-temporal structures more efficiently while maintaining computational feasibility.

\subsection{Spatio-Temporal Prediction} 
Recent advances in spatio-temporal prediction have been driven by the integration of transformer-based models and graph neural networks, aimed at addressing the dual challenges of capturing long-range temporal dependencies and complex spatial interactions. Transformer-based models, such as DLinear~\cite{zeng2023transformers} and TimesNet~\cite{wu2022timesnet}, have demonstrated strong performance in time-series forecasting by leveraging multi-scale temporal patterns and efficient attention mechanisms. Similarly, PatchTST~\cite{nie2022time} introduces patch-based attention to enhance local dependency modeling, while iTransformer~\cite{liu2023itransformer} reconfigures the transformer architecture for multivariate time-series modeling. However, these models often struggle with distribution shifts, such as changes in traffic patterns or external conditions, limiting their robustness in real-world applications.

In the realm of spatial modeling, GNNs and hypergraph structures have emerged as effective tools for capturing complex spatial relationships. STG-NCDE~\cite{choi2022graph} integrates neural controlled differential equations with GNNs to model continuous-time dynamics, while STAEformer~\cite{liu2023staeformer} incorporates spatial and temporal attention mechanisms within a transformer-based framework. Hypergraph-based models, such as STHGCN~\cite{yan2023spatio} and GPT-ST~\cite{li2024gpt}, further advance the field by dynamically capturing higher-order dependencies and spatial drift. However, these approaches frequently rely on static or predefined structures, limiting their ability to adapt to evolving spatial relationships and distribution shifts over time.

A key limitation of existing methods is their inability to effectively model distribution shifts, which are inherent in spatio-temporal systems. For example, STID~\cite{shao2022spatial} simplifies spatio-temporal prediction but lacks adaptability to dynamic spatial interactions, while FEDformer~\cite{zhou2022fedformer} and Autoformer~\cite{wu2021autoformer} primarily focus on temporal modeling without accounting for spatial distribution shifts. These gaps underscore the need for adaptive structures capable of dynamically capturing evolving spatial and temporal patterns. To address these challenges, we propose STH-SepNet, a lightweight framework that leverages adaptive hypergraphs to model distribution shifts and lightweight transformers to capture temporal dynamics, achieving state-of-the-art performance.

\section{Preliminaries}
%\textcolor{red}{DONE} 
\subsection{Problem Formulation}
Given a graph set denoting the spatial feature $ G = (V, E) $, where $V$ and $ E $ represents the set of $ N = |V| $ vertices and the set of edges, respectively. The spatio-temporal prediction problem of multivariate time-series forecasting is defined as follows: suppose the historical observations from $ L $ previous moments $ X_{ (t-L+1):t} \in \mathbb{R}^{L \times N \times F} $, our model STH-SepNet aims to predict the values for the next $ H $ timestamps data $\hat{X}_{ (t+1): (t+H)} \in \mathbb{R}^{H \times N \times F}$. That is, 
\begin{equation}
 \hat{X}_{ (t+1): (t+H)} = \text{STH-SepNet}_{\theta} (X_{ (t-L+1):t}, \hat{A}, \Phi), 
\end{equation}
where $\hat{A}$ is the structural adjacency matrix, $\Phi$ is the prompt of prompt information for input vector, and $\theta$ the parameter of model. 
% In datasets with time-varying network structures, and in the absence of specific instructions, we adopt an adaptive adjacency matrix learning approach by default, enabling the model to effectively capture the underlying temporal dependencies during training.

\subsection{Adaptive Network Construction}
We introduce an adaptive adjacency matrix, $ \tilde{A}_{\text{adp}} $, as input to the ST-Block, aiming to mitigate similarity between adjacent nodes. Given node features $ E_1, E_2 \in \mathbb{R}^{N \times d} $, we employ a shared-parameter feed forward neural network (FFN) to generate node embeddings, which are then mapped to $ F_1, F_2 \in \mathbb{R}^{N \times N} $ as follows: 
\begin{align}
 & F_1 = \tanh (\alpha \mathrm{FFN} (E_1)), \\
 & F_2 = \tanh (\alpha \mathrm{FFN} (E_2)), 
\end{align} 
where $ \alpha $ is a scaling factor that modulates the saturation rate of the activation function. The discrepancy between $ F_1 $ and $ F_2 $ captures directional relationships between nodes. To introduce non-linearity, we construct an asymmetric adjacency matrix $ A_{\text{adp}} \in \mathbb{R}^{N \times N} $: 
\begin{equation}
 A_{\text{adp}} = \mathrm{ReLU} (\tanh (\alpha (F_1^\top F_2 - F_2^\top F_1))).
\end{equation} 
This formulation effectively models asymmetric dependencies by leveraging learned node embeddings in the graph structure.

\subsection{Incident Matrix}
We integrate static spatial topology, e.g., geographic location information in a traffic network, as the static input to build an adjacency matrix $A$, defining node similarity via a negative exponential function of pairwise Euclidean distances. The similarity $ A_{ij} $ between nodes $ i $ and $ j $ is defined as: $
 A_{ij} = \exp\left(-\frac{d_{ij}^{2}}{\sigma^{2}}\right), 
$ 
where $ d_{ij} $ is the distance between nodes $ i $ and $ j $, and $ \sigma $ is a scaling parameter that regulates the effect of distance on similarity. A fixed threshold is applied to maintain the sparsity of the adjacency matrix. 
% Note that other ways to construct the static adjacency matrix may also exist according to the adopted weighting and connecting definition. 

\subsection{Adaptive HyperGraph Construction}
\begin{Definition}\textbf{ (Hypergraph)} 
A high-order graph $H (V, E) $ is defined by a set of $ n $ hypernodes $ V=\{v_1, v_2, \cdots, v_n\} $ and a set of $ m $ hyperedges $ E=\{e_1, e_2, \cdots, e_m\} $, where $ e_j= (v_1^{ (j)}, \cdots, v_{k}^{ (j)}) $ is an unordered set of nodes on hyperedge $ e_j $ , with $ k=|e_j| $ denotes the number of nodes in the hyperedge. 
% When k=1, the higher-order relations degenerate into pair-wise interactions.
\end{Definition}

 \begin{Theorem}
 \label{theorem01}
For any $k \geq 2$, the $ (k-1)$-hops neighborhood of a node $v$, denoted as $N_{k-1} (v)$, corresponds to all nodes involved in the $k$-order hyperedges in $H_v^k$, if and only if the following conditions are satisfied: For each $w\in N_{k-1} (v)$, (1) Local Connectivity Condition: there exists at least one path from $v$ to $w$ consisting of at most $k-1$ hyperedges. 
 (2) Hyperedge Coverage Condition: there exists a $k$-order hyperedge $e\in H_{v}^{k}$ such that $w\in e$ and $e$ contains $v, w$, and at most $k-2$ intermediate nodes. 
 (3) Uniqueness Condition: if there exist multiple $k$-order hyperedges containing both $v$ and $w$, then these hyperedges must share the same set of intermediate nodes.
 Formally:
\begin{equation}
w \in N_{k-1} (v) \iff \{v, F_1, F_2, \ldots, u_k, w\} \in H_v^k, 
\end{equation}
where $F_1, F_2, \ldots, u_k$ are intermediary nodes.
\end{Theorem}

Note that please refer to the Appendix A.1 for detailed proof to the above theorem.
In constructing a $ (k+1)$-order hypergraph from $k-$hop neighborhoods, each node is interconnected with all nodes within its $k$-hops distance, forming one or more hyperedges. 
% Here, a $k-$hop neighborhood comprises nodes reachable from a given node within $k$ edges, capturing higher-order spatial-temporal associations beyond direct connections.
% \textbf{k-hops hypergraphs}.
Given a node $v_i$, its $k$-hops neighborhood $N_{k} (v_i)$ includes all nodes reachable within $k$ edges.
% In particular, when $k=1$ direct neighbors. Along with the increase of $k$, the number of nodes contained in $N_{k} (v_i)$will gradually increase until it covers the whole network.
Similarly, for node features $E_3$, feature representations $F_3$ are obtained via a feedforward neural network:
\begin{equation}
 F_3 = tanh (\alpha FFN (E_3)), 
\end{equation}
where $ \alpha $ controls the activation saturation. Higher-order relationships are then constructed using K-Nearest Neighbors (KNN) on feature representations 
 $ F_3 = [f_1, f_2, \dots, f_n] $. For each node $ v_i $, its nearest $ k $ neighbors $ N (v_i) $ form a hyperedge $ e_i = \{v_i\} \cup N (v_i) $, 
 where $ k = \max_{j} |e_j| $ is the hyperedge order and remains a predefined constant for consistency. 
% 细节在附录A.2.
To determine the hypergraph adptive adjacency matrix $ H_{adp} \in R^{n\times m}$, where $n$ is the number of nodes and $m$ is the number of hyperedges, the adjacency matrix is defined as: $H_{\text{adp}, ij} = 
1 \quad \text{if } v_i \in e_j, \quad \text{and } 0 \quad \text{otherwise}.$
Traditional graph-based methods primarily focus on pairwise interactions between nodes, which can be insufficient for modeling multiple nodes interact simultaneously. By contrast, hypergraphs allow for the representation of higher-order interactions through hyperedges. The adaptive adjacency matrix for hypergraphs can degenerate into traditional graphs, but it leverages this flexibility to capture richer and more complex relationships.

\begin{figure*}[htbp]
 \centering
 \setlength{\abovecaptionskip}{0pt}
 \setlength{\belowcaptionskip}{0pt} 
 \includegraphics[width=0.85\linewidth]{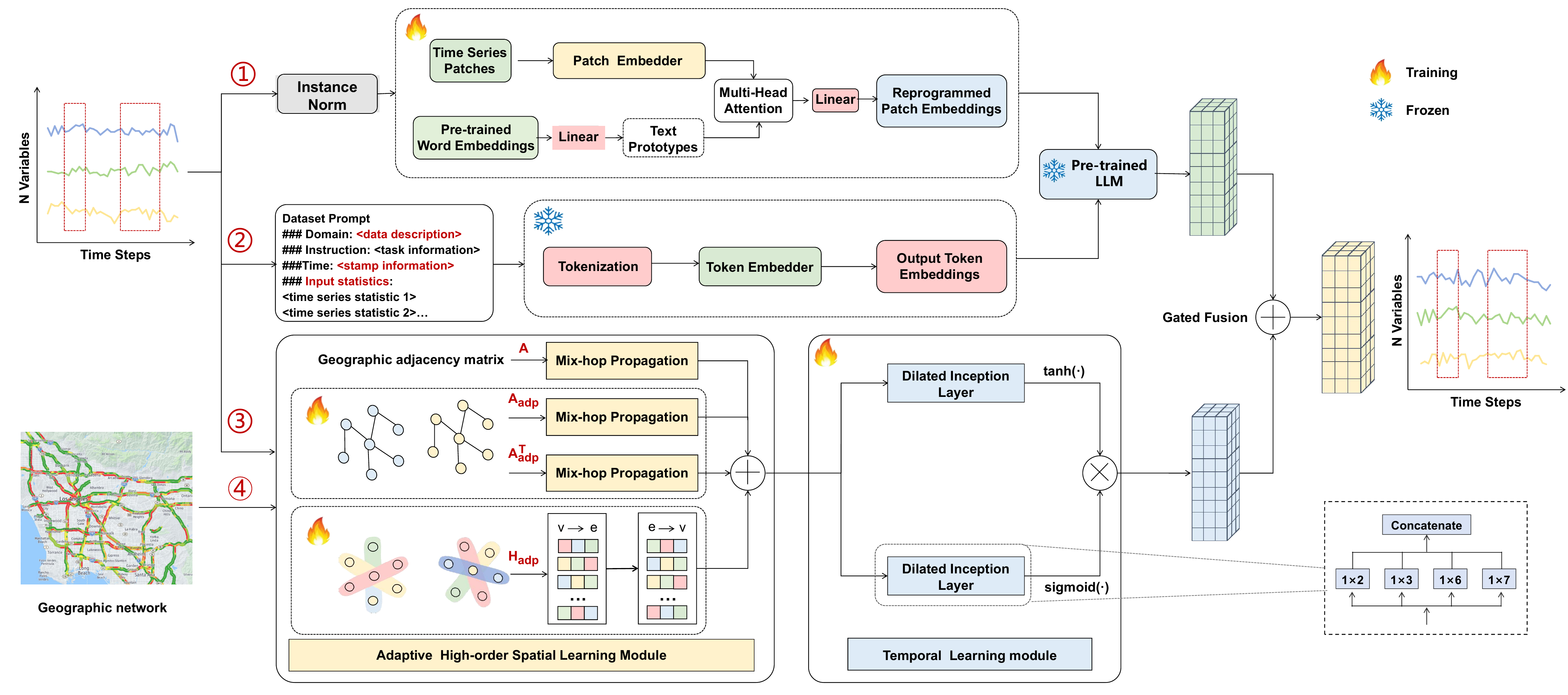}
 \caption{The framework of STH-SepNet. Given a traffic network $G= (V,E)$ and time series $X$ as an example of spatial-temproal datasets. $\bigcirc\!\!\!\!1$ Tokenize and embed $X$ using a customized embedding layer, reprogramming with condensed text prototypes for modality alignment. $\bigcirc\!\!\!\!2$ Incorporate dataset descriptions, task instructions, and statistical characteristics as prompt prefixes to guide input transformation. $\bigcirc\!\!\!\!3$ Leverage a Hypergraph Spatio-Temporal module to model complex spatial dependencies and node-level variations via hierarchical representation learning.  $\bigcirc\!\!\!\!4$ Incident matrix: real geographic network, if not, Adaptive Graph or Adaptive HyperGraph is used.
 By integrating $\bigcirc\!\!\!\!1$ $\bigcirc\!\!\!\!2$ $\bigcirc\!\!\!\!3$ , STH-SepNet generate the forecasts.}
 \label{fig02}
\end{figure*}

\section{Methodology}
In this work, we propose STH-SepNet, a spatio-temporal forecasting model that integrates a pre-trained LLM with adaptive hypergraphs. 
% This design captures global trends via LLMs and local heterogeneity through HGNNs.
As shown in Figure~\ref{fig02}, STH-SepNet comprises two key components: lightweight large language models for
temporal dynamics and adaptive hypergraphs for spatial dependencies. 
% The global trend extraction module uses LLMs with statistical prompt design to predict future trends, aligning time series data with textual representations for enhanced forecasting. The hypergraph spatio-temporal module constructs an adaptive hypergraph to model complex spatial dependencies and node-level variations, refining predictions through hierarchical representation learning. By integrating LLM-driven trend modeling with HGNN-based spatial-temporal learning, STH-SepNet effectively captures both macroscopic trends and fine-grained heterogeneity, leading to improved predictive performance.

\subsection{Global Trend Module}
\textbf{Local Aggregation Module.} This module processes the model input node features $ X \in \mathbb{R}^{B \times N \times T \times F} $ by executing an average pooling operation to extract the common features of all nodes within a region, capturing the overall fluctuation trends. That is, 
\begin{equation}
\label{avgpool}
X_{\text{pool}} = \text{AvgPool} (X), 
\end{equation}
where $ X_{\text{pool}} \in \mathbb{R}^{B \times T \times F} $, $ B $ is the batch size, $ T $ denotes the time steps, and $ F $ equals 1. As shown in Figure~\ref{fig02} $\bigcirc\!\!\!\!1$ , the time series embedding module reduces the computational time and memory complexity of the model by aggregating information across adjacent time steps and utilizing temporal patches \cite{nie2022time}. Specifically, $ X_{\text{pool}} $ is partitioned into overlapping or non-overlapping blocks $ X_P \in \mathbb{R}^{P \times N_P} $, where $ P $ is the window length and the number of sliding windows is computed as $N_P = (T - P) / S + 2 $, with the stride $ S $. Each patch is treated as a time series token and embedded to obtain $ \hat{X}_P \in \mathbb{R}^{P \times d_m} $, where $d_m$ is the hidden dimension of the LLM.
 
\vspace{-5pt}
\subsection{Prompt Adaptation Module} 
Given that LLMs are primarily trained on extensive text corpora and lack inherent time series knowledge, we propose a cross-modal alignment strategy that transforms time-series data into textual tokens, enabling LLMs to leverage their reasoning capabilities for specialized forecasting tasks. To enhance predictive accuracy, we adopt Pattern-Exploiting Training , which formulates natural language templates as prompts within the embedding space. These prompts integrate three key components: dataset description, task instructions, and statistical characteristics. More details refer to Appendix B.2. 
We further prepend this structured information as a prefix prompt, concatenating it with aligned temporal embeddings before feeding it into the LLM. This enables the model to generate valid outputs and adapt to downstream tasks. 

\textbf{LLM Module.} We utilize a partially frozen pre-trained LLM to capture temporal dependencies in traffic data, fine-tuning its feed-forward layers via LoRA \cite{hu2021lora}. Pretrained models serves as the backbone for STH-SepNet, comprising stacked transformer decoder modules with $N$ layers.
As depicted in Figure.~\ref{fig02}~$\bigcirc\!\!\!\!2$ , the input to each layer is represented $ z = \{z^1, z^2, \dots, z^N\} $, where $ z^1 $ consists of concatenated prompt and time-series embeddings
For the $ i $-th layer, the input $ z^i $ undergoes multi-head self-attention (MHSA) followed by layer normalization (LN), producing an intermediate state $ \tilde{z}^i $, which is further processed by a feed-forward network and another layer normalization step to yield $z^{i+1}$ for the next layer. It can be summarized as:
\begin{align}
 & (Q_i, K_i, V_i) =(W^Q_i z^i, W^K_i z^i, W^V_i z^i), \\
 & \text{head}_i = \text{softmax}\left(\frac{Q_i K_i^T}{\sqrt{d}}\right) V_i, \\
 & \text{MHSA}(z^i, z^i, z^i) = W(\text{head}_1 \| \ldots \| \text{head}_h), \\
 & \tilde{z}^i = \text{LN}(z^i + \text{MHSA}(z^i, z^i, z^i)), \\
 & z^{i+1} = \text{LN}(\tilde{z}^i + \text{FFN}(\tilde{z}^i)), 
% & \text{FFN} (\tilde{z}^i) = \text{ReLU} (W_1 \tilde{z}^i + b_1) W_2 + b_2.
\end{align}
where $ z^i $ is the input hidden state at the $ i $-th layer, and $ \tilde{z}^i $ is the intermediate state after MHSA and layer normalization. Since the LLM module outputs token sequences, we apply a linear layer to align learned patch representations. Additionally, pretrained models such as BERT \cite{devlin2018bert}, GPT2, LLAMA \cite{touvron2023llama} and Deepseek \cite{yang2024qwen25,guo2025deepseek} can be employed for autoregressive token prediction.

\subsection{Hypergraph Spatio-temporal Module}
To enhances spatio-temporal prediction by incorporating higher-order coupling relationships, STH-SepNet consists of four key components: Mixed Multi-Layer Information Aggregation Module, Adaptive Graph Convolution Network, Hypergraph Convolutional Network, and Temporal Convolutional Network in Figure.~\ref{fig02} $\bigcirc\!\!\!\!3$ .

\textbf{Mixed Multi-Layer Information Aggregation Module} (MixProp).
% This module integrates multi-layer neighborhood propagation, residual connections, and a gating mechanism to enhance information transmission and stability. 
Given an input $ X^{ (0)}\in R^{N\times C}$, where 
$N$ represents the number of nodes and $C$ is the feature dimension, the module updates node features through $ k $-layer propagation. The feature propagation at layer $ k+1 $ is formulated as, 
\begin{equation}
 X^{ (k+1)} = \alpha * X^{ (k)} + (1-\alpha)\hat{A} X^{ (k)}, 
\end{equation}
where $X^{ (k)}$ is the node feature matrix at the $k$-th layer, $ \hat{A} = D^{-\frac{1}{2}} (A+I)D^{-\frac{1}{2}}$ is the normalized adjacency matrix, $D$ is the degree matrix, $I$ is the identity matrix, $\alpha$ controls the weight of residual connections. 
To further regulate the flow of information, the MixProp module incorporates a gating mechanism defined by:
\begin{equation}
\centering
\begin{aligned}
 & G^{ (k)} = \sigma (W_gX^{ (k)}), \\
 & X^{ (k+1)} = G^{ (k)} \odot X^{ (k)} + (1- G^{ (k)})\odot \hat{A} X^{ (k)},
\end{aligned}
\end{equation}
where $W_g\in R^{C\times C}$, $\sigma (\cdot)$ is the sigmoid activation function, and $\odot$ denotes element-wise multiplication. The MixProp module employs k-layer propagation to expand the receptive field and capture dependencies between distant nodes.

\textbf{Adaptive Graph Convolution Network Module.}
To capture both structural and adaptive spatial dependencies, we apply three MixProp-based graph convolutions on the adaptive adjacency matrix $A_{adp}$ and the real road network $A$. These operations extract first-order, transposed first-order, and real relationships, defined as:
\begin{align}
 & X_{gconv1} = \text{MixProp} (X, A_{adp}, K, \alpha), \\
 & X_{gconv2} = \text{MixProp} (X, A^{T}_{adp}, K, \alpha), \\
 & X_{gconv3} = \text{MixProp} (X, A, K, \alpha), 
\end{align}
where $X\in R^{B\times N \times T}$ represents the input time-series features, $K$ is the number of propagation layers, and $\alpha$ controls residual weighting. The final spatial representation is obtained by fusing these outputs:
\begin{equation}
X_{GCN} = X_{gconv1} + X_{gconv2} + X_{gconv3}.
\end{equation}
% This design enables robust multi-scale feature extraction, improving adaptability to evolving spatial structures.

\textbf{Adaptive Hypergraph Convolution Network Module.}
Given an input feature matrix $X\in R^{B\times N\times T}$ and an adaptive hypergraph adjacency matrix $H_{adp}$, this module employs two information aggregation mechanisms: from node to hyperedge and from hyperedge to node.
Initially, a feedforward neural network (FFN) transforms the feature matrix. In the node-to-hyperedge process, each hyperedge $e_j$
 accumulates information from its associated nodes $N (e_j)$, ollowed by hyperedge aggregation and linear transformation: 
\begin{align}
\label{eq19}
 & X^{enc} = FFN(X), \\
 & X^{enc}_{e} = \sigma \left( \sum_{i\in N(e)} H_{adp, i} X_{i}^{enc} W \right),
\end{align}
where $W$ represents the trainable parameter matrix, and $\sigma (\cdot)$ is the ReLU activation function. Subsequently, features of all hyperedges containing a node $v_i$ are aggregated back to node representation, 
\begin{align}
\label{eq21}
 X_{v}^{enc} = \sum_{j \in \mathcal{E}(v_i)} H_{adp,j} X_{e_j}^{enc}, 
\end{align}
where $\mathcal{E}(v_i) $ indicates the set of hyperedges for node $v_i$.
The output of the hypergraph spatial learning module is $X_ {HGCN} = X_{v}^ {enc}$.

To integrate pairwise and high-order spatial features, we fuse representations from GCNs and HGCNs, ensuring strong scalability in the spatial module:

\begin{equation}
 X = \gamma X_{GCN} + (1- \gamma) X_{HGCN}, \gamma \in [0, 1], 
\end{equation}
where $\gamma$ is a tunable parameter. When $\gamma=1$, the model degenerates into standard spatial learning module, whereas $\gamma=0$ captures only higher-order dependencies.

\textbf{Spatial-temporal Convolution Module.} The Spatio-Temporal Convolution Module consists of multiple stacked ST-Blocks, each containing an S-Block for spatial dependencies and a T-Block for temporal dependencies.
In the S-Block, each node state $ h^{ (v)}_{t} $ is initialized as the input features $ X \in \mathbb{R}^{B \times N \times T \times F} $ 
and updated by aggregating features from its neighbors, as follows:
\begin{align}
 & m_t^{v} = \sum_{u \in N (v)} h_{t-1}^{u} , \\
 & h_t^{v} = \sigma ( (1 + \epsilon) h_{t-1}^{v} + m_t^{v}) , 
\end{align}
where $ m_t^{v} $ is the aggregated neighborhood feature at time $ t $, $ \sigma$ is an activation function and $ \epsilon $ is a learnable parameter.

The T-Block comprises 1-D dilated convolution layers with a gating mechanism featuring only an output gate. 
Given the input $\chi \in \mathbb{R}^{T \times N \times F}$, the gated output $h$ is defined as:

\begin{equation}
h = \tanh (q (\chi)) \odot \sigma (q (\chi)) , 
\end{equation}
where $q (\chi)$ is the output of the dilated convolution layers, $\odot$ denotes the Hadamard product, and $\sigma$ is the sigmoid activation function.

\begin{table*}[htbp]
 \centering
 \setlength{\abovecaptionskip}{0pt}
 \setlength{\belowcaptionskip}{0pt} 
\caption{Performance comparison. Multivariate forecasting results with a prediction horizon of 48 time steps and a fixed lookback window of T=48. Bolded results indicate the best performance. (LLMs: BERT)}
\label{table01}
 \renewcommand{\arraystretch}{0.95} % 调整行距
 \setlength{\tabcolsep}{8.5pt} % 调整列间距
\begin{tabular}{lllllllllll}
\hline
\hline
 Model & \multicolumn{2}{c}{BIKE-Inflow} & \multicolumn{2}{c}{BIKE-Outflow} & \multicolumn{2}{c}{PEMS03} & \multicolumn{2}{c}{BJ500} & \multicolumn{2}{c}{METR-LA}\\
 \cmidrule (lr){2-3} \cmidrule (lr){4-5} \cmidrule (lr){6-7}\cmidrule (lr){8-9} \cmidrule (lr){10-11}
 & MAE & RMSE & MAE & RMSE & MAE & RMSE & MAE & RMSE & MAE & RMSE \\
 \midrule
 Autoformer (NIPS, 2021)\cite{wu2021autoformer} & 7.01 & 17.52 & 7.19 & 17.75 & 44.87 & 70.84 & 10.79 & 16.06 & 12.47 & 20.04 \\
 Informer (AAAI, 2021)\cite{zhou2021informer} & 8.25 & 20.37 & 9.21 & 21.50 & 33.72 & 52.15 & 7.58 & 11.96 & 14.50 & 20.35\\
 FEDformer (PMLR, 2022)\cite{zhou2022fedformer} & 6.28 & 16.30 & 6.56 & 16.67 & 35.00 & 50.84 & 10.77 & 15.99 & 12.35 & 18.79 \\
 DLinear (AAAI, 2023) \cite{zeng2023transformers} & 5.71 & 15.49 & 5.82 & 15.36 & 45.30 & 66.81 & 8.55 & 13.49 & 10.90 & 17.31 \\
 TimesNet (ICLR, 2023)\cite{wu2023timesnet} & 5.54 & 15.41 & \underline{5.56} & \underline{15.18} & 37.54 & 62.99 & 8.67 & 13.96 & 10.22 & 18.29\\ 
 PatchTST (ICLR, 2023)\cite{nietime} & \underline{5.53} & \underline{15.39} & 5.63 & 15.23 & 48.42 & 78.24 & 8.79 & 14.28 & \underline{10.13} & \underline{18.27} \\
 iTransformer (ICLR, 2024)\cite{liu2023itransformer} & 6.05 & 16.39 & 6.15 & 16.69 & 43.63 & 70.61 & 9.01 & 14.32 & 10.15 & 18.36\\
 TIMELLM (ICLR, 2024)\cite{jin2024time} & 6.81 & 16.72 & 6.93 & 16.30 & \underline{32.62} & \underline{49.77} & \underline{7.25} & \underline{11.58} & 12.36 & 18.53 \\
 AdaMSHyper (NIPS, 2024)\cite{shangada} & 6.72 & 16.91 & 7.04 & 17.14 & 33.49 & 50.37 & 7.41 & 11.60 & 12.51 & 18.60\\
 \midrule
 AGCRN (NIPS, 2020)\cite{bai2020adaptive} & 6.64 & 16.14 & 6.77 & 16.36 & 33.14 & 54.88 & 6.32 & 12.81 & 11.39 & 23.15 \\
 ASTGCN (AAAI, 2019)\cite{guo2019attention} & 6.66 & 15.87 & 6.26 & 14.48 & 30.65 & 53.96 & 6.34 & 11.34 & 10.54 & 22.76\\
 MSTGCN (TNSRE, 2021)\cite{jia2021multi} & \underline{5.91} & \underline{14.11} & 6.04 & 14.24 & 29.57 & 47.97 & \underline{5.62} & \underline{11.15} & 10.17 & 20.24\\
 MTGNN (KDD, 2020)\cite{wu2020connecting} & 6.16 & 14.80 & \underline{5.93} & \underline{13.93} & \underline{29.04 } & \underline{50.32} & 5.86 & 10.91 & 9.98 & 21.23\\
 STGODE (KDD, 2021)\cite{fang2021spatial} & 6.77 & 15.93 & 6.82 & 15.50 & 33.39 & 54.16 & 6.44 & 12.14 & 11.48 & 22.85\\
 STSGCN (AAAI, 2020)\cite{song2020spatial} & 6.73 & 15.89 & 6.58 & 15.36 & 34.23 & 58.07 & 6.40 & 12.03 & 11.07 & 22.79\\
 STGCN (IJCAI, 2018)\cite{yu2018spatio} & 7.08 & 15.72 & 7.36 & 16.11 & 36.02 & 53.44 & 6.73 & 12.62 & 12.38 & 22.55 \\
 % TGCN (2019)\cite{zhao2019t} & 7.61 & 17.15 & 7.91 & 17.19 & 38.72 & 58.29 & 7.24 & 13.46 & 13.31 & 24.59\\
 GMAN (AAAI, 2020)\cite{zheng2020gman} & 6.73 & 15.60 & 6.94 & 15.84 & 33.96 & 53.02 & 6.41 & 12.40 & 11.69 & 22.37\\
 STAEformer (CIKM, 2024)\cite{liu2023staeformer} & 5.97 & 14.57 & 6.17 & 14.70 & 29.62 & 48.03 & 5.79 & 10.42 & \underline{9.91} & \underline{21.17} \\
 STD-MAE (IJCAI, 2024) \cite{gao2024spatial} & 6.13 & 14.87 & 6.21 & 14.37 & 30.40 & 48.38 & 5.92 & 11.49 & 10.52 & 23.11\\
 \midrule
 % STH-SepNet-GNN (Ours) & \textbf{5.03} & \textbf{13.29} & 5.47 & 14.36 & 21.11 & 34.52 & \textbf{5.57} & \underline{9.53} & 9.44 & 17.35 \\
 STH-SepNet (Ours) & \textbf{5.18} & \textbf{14.40} & \textbf{5.33} & \textbf{14.23} & \textbf{21.03} & \textbf{34.17} & \textbf{5.58} & \textbf{9.77} & \textbf{9.42} & \textbf{16.41} \\
 % STH-SepNet-MixGNN (Ours) & \underline{5.06} & \underline{13.72} & \underline{5.39} & \underline{13.97} & \underline{21.09} & \underline{34.39} & 5.61 & \textbf{9.07} & \textbf{9.35} & \underline{16.48}
 % \\ 
 \hline\hline
\end{tabular}
\end{table*}

\subsection{Gated Fusion Module}
To integrate global trends and node heterogeneity, we fuse outputs from the pre-trained LLM and adaptive high-order spatial module, denoted as $ O_1, O_2 \in \mathbb{R}^{B \times T \times N}$. A feedforward neural network (FFN) maps the concatenated output vectors to gates of equivalent dimensions. The gating process is formulated as: 
\begin{equation}
 \text{Gate} = \sigma (\mathrm{FFN} ([O_1, O_2])), 
\end{equation}
where $\sigma$ represents the sigmoid activation function, and $[\cdot, \cdot]$ denotes the concatenation operation. The ultimate gated fusion process can be expressed as:
\begin{equation}
 \tilde{O} = O_1 \odot \text{Gate} + O_2 \odot (1 - \text{Gate}), 
\end{equation}
where $\text{Gate}$ signifies the gate map, and $\tilde{O} \in \mathbb{R}^{B \times T \times N}$ represents the resultant fused output.

\section{Experiments}

To verify the effectiveness and performance of the STH-SepNet model, we address the following questions:
\textbf{RQ1:} How does our proposed STH-SepNet perform on datasets compared with state-of-the-art baselines? 
\textbf{RQ2:} Does adaptive graph convolution, particularly adaptive higher-order graph convolution, enhance predictive performance over static graph convolution for large models?
\textbf{RQ3:} Whether the large model prediction structure is a necessary component of the proposed model?
\textbf{RQ4:} How do large model parameters affect prediction performance?

\subsection{Experiment Settings} 
\textbf{Datasets.} We conduct the experiments on five datasets: BIKE-Inflow, BIKE-Outflow, PEMS03, BJ500 and METR-LA~\cite{li2023dynamic}. where partition the datasets into train/validation/test sets by the ratio of 7:1:2. Appendix B.1 contains more dataset details. 
 \\
\textbf{Baselines.} As aforementioned, STH-SepNet is designed as a general framework for spatio-temporal prediction tasks. To evaluate its effectiveness, we compare it with several state-of-the-art time series prediction models, including Autoformer~\cite{wu2021autoformer}, 
Informer~\cite{zhou2021informer}, FEDformer~\cite{zhou2022fedformer}, 
DLinear~ \cite{zeng2023transformers}, TimesNet~\cite{wu2023timesnet}, 
PatchTST~\cite{nietime}, iTransformer~\cite{liu2023itransformer}, TIMELLM~\cite{jin2024time}, AdaMSHyper~\cite{shangada}. Additionally, to demonstrate the robustness of our model, we also include comparisons with spatio-temporal prediction models such as AGCRN~\cite{bai2020adaptive}, MSTGCN~\cite{jia2021multi}, MTGNN~\cite{wu2020connecting}, STGODE~\cite{fang2021spatial} , STSGCN~\cite{song2020spatial}, STGCN~\cite{yu2018spatio}, GMAN~\cite{zheng2020gman}, STAEformer~\cite{liu2023staeformer}, STD-MAE~\cite{gao2024spatial}.
\subsection{Main Results}
\subsubsection{\textbf{Effectiveness of STH-SepNet. (RQ1)}}
Table \ref{table01} presents the performance of the proposed STH-SepNet method integrating pretrained model BERT across five datasets . Our method achieves the best Mean Absolute Error (MAE) and Root Mean Squared Error (RMSE) results, which can be attributed to its spatio-temporal separation strategy and adaptive hypergraph structure. Key advantages are detailed below:\\
\textbf{Decoupled Spatio-Temporal Modeling.} STH-SepNet addresses the limitations of joint spatio-temporal modeling by isolating temporal and spatial dependencies. On the BIKE-Outflow dataset, where non-stationary spatial events (e.g., weather-induced station closures) intersect with periodic temporal trends (e.g., rush hours), our method achieves an MAE of 5.33 and RMSE of 14.23, outperforming joint modeling frameworks like TimesNet (MAE: 5.56) and PatchTST (MAE: 5.63). By decoupling temporal dynamics (handled by lightweight modules) and spatial dependencies (modeled via adaptive hypergraphs), we mitigate interference between heterogeneous features, ensuring robust predictions in dynamic scenarios. \\
\textbf{Adaptive Hypergraphs for Dynamic Spatial Drift.}
We propose an adaptive hypergraph structure to model evolving spatial dependencies. On the PEMS03 dataset, where traffic accidents or construction events disrupt road network interactions, our method reduces RMSE to 34.17, a 28.8\% improvement over dynamic graph-based approaches like STAEformer (RMSE: 48.03). The dynamic hyperedge generation mechanism allows our method to adjust node relationships dynamically, capturing spatial drift. For example, on the BJ500 dataset, our method achieves an MAE of 5.58, surpassing MTGNN (MAE: 5.86) and MSTGCN (MAE: 5.62), demonstrating its ability to adapt to sudden changes in spatial dependencies. \\
\textbf{Scalable Efficiency.} Our approach significantly reduces computational complexity by decoupling the node dimension, transforming spatio-temporal prediction into parallelizable univariate tasks. On the METR-LA dataset (large-scale road network), our method achieves an MAE of 9.42 and RMSE of 16.41, outperforming both graph-based models (MTGNN: MAE: 9.98) and transformer variants (iTransformer: MAE: 10.15). Compared to large language model (LLM)-based baselines like TIMELLM (MAE: 12.36 on METR-LA), our method reduces MAE by 23.8\%. This efficiency is achieved through lightweight LLM adaptations for temporal modeling and node-wise decoupling, avoiding the parameter bloat of monolithic LLM frameworks while maintaining high accuracy.
 \vspace{-5pt}
\subsubsection{\textbf{Effectiveness of Adaptive Hypergraph Structure. (RQ2)}}
Table \ref{tab:model_comparison_rubost} presents the performance of the proposed method using different graph representation strategies, including a static graph (STH-SepNet-static), an adaptive graph convolutional network (STH-SepNet-GNN), and an adaptive hypergraph structure (STH-SepNet). Additionally, we compare these methods with the baseline large model TimeLLM. The results indicate that STH-SepNet achieves the best performance, and the effectiveness of the adaptive hypergraph can be attributed to the following factors:\\
\textbf{Dynamic Adaptation to Complex Spatial Dependencies.} 
The core strength of the adaptive hypergraph lies in its dynamic adjustment capability, which enables real-time responses to evolving spatial dependencies. In transportation networks, traditional static graphs or GNNs rely on predefined adjacency relationships and fail to capture sudden disruptions (e.g., traffic accidents) that alter node correlations. On the BIKE-Outflow dataset, STH-SepNet notably reduces prediction errors (MAE: 5.33 vs. 6.34 for the static variant) by dynamically generating hyperedge weights. This adaptability allows the model to flexibly represent shifts in regional traffic flows, such as adjusting inter-regional influence weights under policy-driven restrictions (e.g., traffic bans). Static graphs, constrained by fixed structures, cannot accommodate such changes. This capability is critical in dynamic scenarios (e.g., abrupt traffic flow shifts), enhancing the model’s efficiency in capturing spatial drift.  
\begin{table}[htbp]
 \centering
 \small
 \setlength{\abovecaptionskip}{0pt}
 \setlength{\belowcaptionskip}{0pt} 
 \caption{Comparison with different LLMs of adaptive high-order and low-order spatio-temporal multitime series forecasting results.}
 \label{tab:model_comparison_rubost}
 \begin{tabular}{llcccc}
 \hline
 \hline
 \multicolumn{2}{c}{\multirow{2}{*}{Model}} & \multicolumn{2}{c}{BIKE-Outflow } & \multicolumn{2}{c}{PEMS03 } \\
 \cmidrule (lr){3-4} \cmidrule (lr){5-6}
 \multicolumn{2}{c}{} & MAE & RMSE & MAE & RMSE \\
 \midrule
 \multirow{4}{*}{BERT} & TIMELLM & 6.74 & 16.13 & 32.68 & 50.39 \\ 
 & STH-SepNet-Static & 6.34 & 16.41 & 29.53 & 48.94 \\ & STH-SepNet-GNN & \underline{5.47} & \underline{14.36} & \underline{21.11} & \underline{34.52} \\
 & STH-SepNet & \textbf{5.33} & \textbf{14.23} & \textbf{21.03} & \textbf{34.17} \\
 \midrule
 \multirow{4}{*}{GPT2} 
 & TIMELLM & 6.93 & 16.30 & 32.63 & 49.83 \\ 
 & STH-SepNet-Static & 6.53 & 16.61 & 30.03 & 49.07 \\
 & STH-SepNet-GNN & \underline{5.68} & \underline{14.48} & \underline{21.85} & \underline{35.78} \\
 & STH-SepNet & \textbf{5.31} & \textbf{14.24} & \textbf{21.43} & \textbf{35.01} \\
 \midrule
 \multirow{4}{*}{GPT3} & TIMELLM & 7.01 & 16.63 & 34.85 & 51.89 \\ 
 & STH-SepNet-Static & 6.79 & 16.71 & 30.64 & 49.16 \\
 & STH-SepNet-GNN & \underline{5.66} & \underline{13.90} & \underline{21.27} & \underline{34.78} \\
 & STH-SepNet & \textbf{5.24} & \textbf{14.16} & \textbf{21.13} & \textbf{34.69} \\
 \midrule
 \multirow{4}{*}{LLAMA1B} & TIMELLM & 7.10 & 16.74 & 34.06 & 51.59 \\ 
 & STH-SepNet-Static & 6.27 & 16.40 & 29.76 & 48.63 \\
 & STH-SepNet-GNN & \underline{5.87} & \underline{14.80} & \underline{22.19} & \underline{35.87} \\
 & STH-SepNet & \textbf{5.29} & \textbf{14.20} & \textbf{21.37} & \textbf{34.92} \\
 \midrule
 \multirow{4}{*}{LLAMA7B} & TIMELLM & 6.95 & 16.41 & 34.17 & 52.47 \\ 
 & STH-SepNet-Static & 6.73 & 16.87 & 30.24 & 49.92 \\
 & STH-SepNet-GNN & \underline{5.91} & \underline{14.94} & \underline{21.64} & \underline{35.24} \\
 & STH-SepNet & \textbf{5.34} & \textbf{14.31} & \textbf{21.52} & \textbf{35.17} \\
 \midrule
 \multirow{4}{*}{LLAMA8B} & TIMELLM & 7.02 & 16.55 & 35.72 & 51.97 \\ 
 & STH-SepNet-Static & 6.85 & 16.64 & 30.47 & 49.98 \\
 & STH-SepNet-GNN & \underline{5.70} & \underline{14.27} & \underline{21.57} & \underline{35.23} \\
 & STH-SepNet & \textbf{5.28} & \textbf{14.20} & \textbf{21.51} & \textbf{35.19} \\
\midrule
 \multirow{4}{*}{DeepSeek1.5B} & TIMELLM & 6.94 & 16.25 & 33.19 & 50.28\\ 
 & STH-SepNet-Static & 6.79 & 16.37 & 30.26 & 48.81 \\
 & STH-SepNet-GNN & \underline{5.74} & \underline{14.55} & \underline{21.73} & \underline{35.47} \\
 & STH-SepNet & \textbf{5.27} & \textbf{14.19} & \textbf{21.39} & \textbf{34.96} \\
 \hline
 \hline
 \end{tabular}
 \vspace{-10pt}
\end{table}
\\
\textbf{Enhancing Lightweight Models to Surpass Large Counterparts.}
While large models like TIMELLM leverage massive parameters to capture complex patterns, their performance in spatio-temporal prediction is outperformed by lightweight LLMs integrated with adaptive hypergraphs. For example, on the PEMS03 dataset, STH-SepNet with a BERT backbone achieves an RMSE of 34.17, far superior to TIMELLM’s 50.39. This highlights the division-of-labor advantage of the adaptive hypergraph: it specializes in modeling spatial dynamics, while the lightweight LLM focuses solely on single-node temporal trends, avoiding feature interference inherent in joint modeling. In traffic sensor networks, the hypergraph independently models cascading effects of multi-road congestion, while the LLM predicts traffic trends for individual sensors. This decoupling reduces model complexity and improves task-specific focus, enabling lightweight models to outperform large counterparts even in resource-constrained scenarios. \\
\textbf{Cross-Architecture Consistency and Higher-Order Interaction Modeling.}
The adaptive hypergraph’s design is architecture-agnostic, delivering consistent performance across diverse LLM backbones (BERT, GPT, LLAMA, Deepseek). For instance, with the LLAMA7B backbone, STH-SepNet achieves an RMSE of 35.17 on PEMS03, outperforming both the GNN variant (RMSE: 35.24) and static version (RMSE: 49.92). This consistency stems from the hypergraph’s explicit modeling of higher-order spatial interactions. Traditional GNNs capture only pairwise node interactions, whereas hypergraphs connect multiple nodes via hyperedges, simultaneously modeling multi-region joint influences (e.g., concurrent congestion across multiple roads amplifying impacts on central areas). In large-scale road networks like METR-LA, this capability enables accurate prediction of congestion propagation paths, while static graphs or standard GNNs exhibit significant biases due to their inability to represent multi-node effects. The explicit modeling of higher-order interactions distinguishes adaptive hypergraphs as a uniquely powerful tool for spatial dependency learning. 

\subsection{Ablation Study}
We conduct three ablation studies to validate the effectiveness of key components in our proposed framework, exploring the role of LLMs, the mixed-order spatio-temporal convolutional networks and the order of hypergraph on forecast results.
 \vspace{-8pt}
\subsubsection{\textbf{LLMs play a critical role in temporal modeling. (RQ3)}}
To assess the necessity of LLMs in spatio-temporal prediction and their synergistic effects with the adaptive hypergraph structure, we conduct systematic ablation studies. 
Specifically, we compare a model variant without LLMs (w/o) only 
the adaptive hypergraph and linear temporal layers, against the full framework LLMs (STH-SepNet) equipped with different LLM backbones, such as BERT (w/i), GPT (w/i), LLAMA (w/i) and Deepseek (w/i).

The comparison between STH-SepNet-w/o and its LLM-enhanced counterparts reveals substantial performance improvements, particularly in capturing long-range dependencies and multi-scale periodicity. As illustrated in Figure~\ref{fig03},
on the BIKE-Outflow dataset, the LLM-enhanced variant (BERT-w/i) significantly outperforms the non-enhanced version, and the integration of an LLM into STH-SepNet-GNN markedly boosts prediction accuracy. This enhancement can be attributed to the LLM's capability to learn rich statistical characteristics from batch inputs, including minimum, maximum, median, and trend information. Consequently, the model effectively extracts semantic temporal features and complex periodic trends from traffic spatio-temporal data. 

\subsubsection{\textbf{Lightweight LLMs achieve competitive performance while maintaining architectural stability, highlighting the efficiency of the proposed framework. (RQ4)} }
\begin{figure}[htbp]
 \centering
 \setlength{\abovecaptionskip}{0pt}
 \setlength{\belowcaptionskip}{-5pt} 
 \includegraphics[width=1\linewidth]{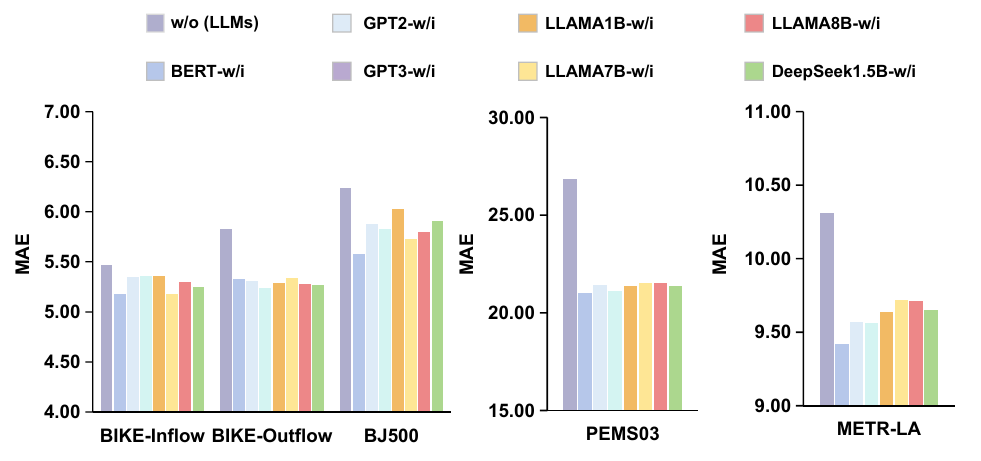}
 \caption{Performance comparison of MAE between STH-SepNet trained on different datasets.}
 \label{fig03}
\end{figure}

As shown in Figure~\ref{fig03}, on the STH-SepNet models, incorporating a BERT model demonstrates notable accuracy. For example,
on the BIKE-Inflow dataset, BERT-w/i not only surpasses larger backbone models such as LLAMA7B-w/i but does so with fewer parameters (BERT:110M, LLAMA7B:6740M), indicating that excessive parameter scaling is not essential for effective temporal modeling.
On the BIKE-Inflow dataset, the BERT-w/i not only surpasses larger backbones such as LLAMA7B-w/i but also operates with less than 10\% fewer parameters, demonstrating that excessive parameter scaling is not a prerequisite for effective temporal modeling. 
Moreover, results on the BJ500 dataset indicate that performance variations across different LLM backbones remain minimal, with fluctuations under 3\%. This stability arises from the decoupled nature of the adaptive hypergraph, which independently captures spatial dependencies, thereby reducing reliance on LLM scale and ensuring that lightweight models remain competitive. We also provide a details comparison in Appendix B.3.

\begin{figure}[htbp]
 \centering
 \setlength{\abovecaptionskip}{0pt}
 \setlength{\belowcaptionskip}{-5pt} 
 \includegraphics[width=1\linewidth]{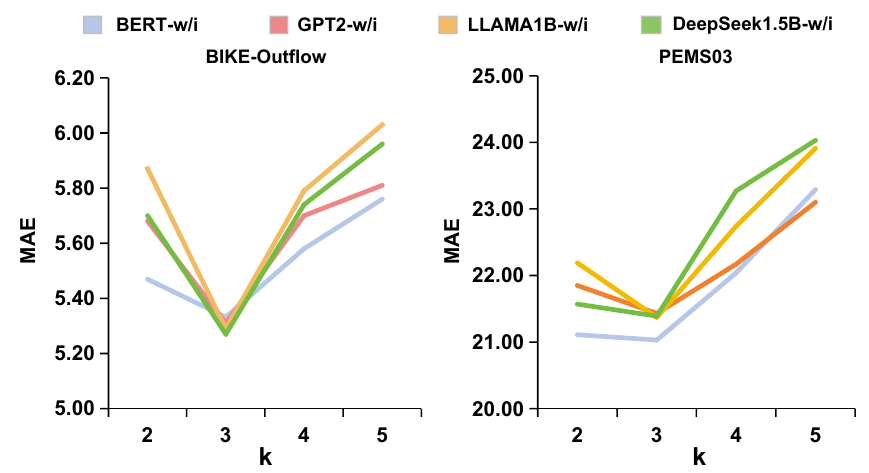}
 \caption{Analysis of effective order on adaptive hypergraph. }
 \label{fig04}
\end{figure}

\begin{figure}[htbp]
 \centering
  \setlength{\abovecaptionskip}{0pt}
 \setlength{\belowcaptionskip}{-5pt} 
 \includegraphics[width=1\linewidth]{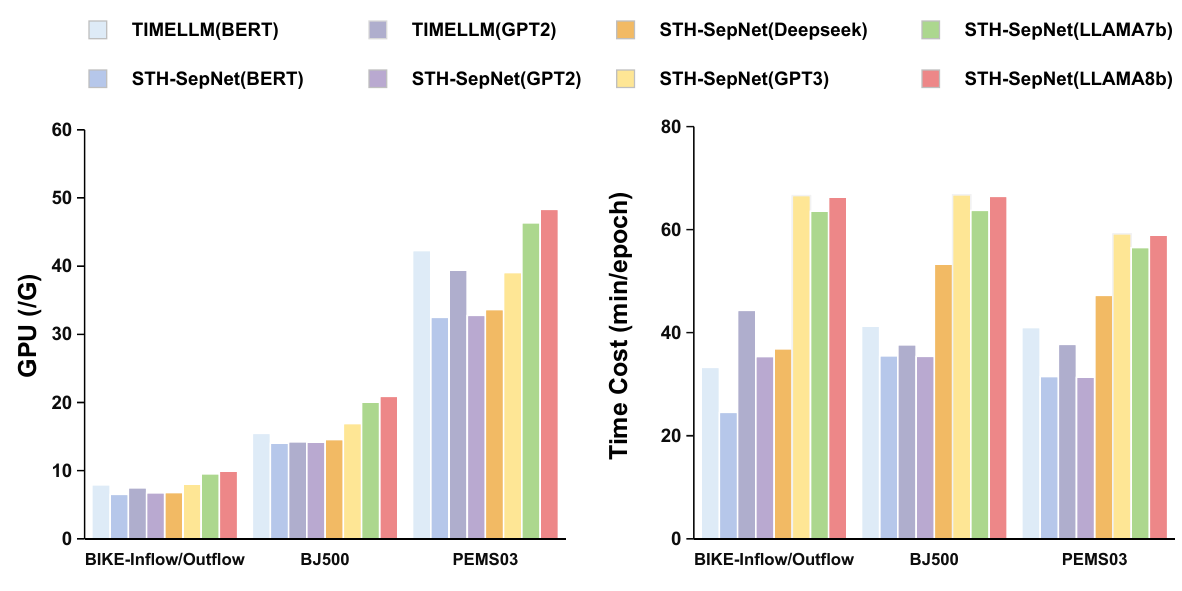}
 \caption{Comparison of GPU and time complexity. }
 \label{fig05}
\end{figure}

\subsubsection{\textbf{The effective order of the hypergraph can enhance the performance of the model.} }
Our proposed framework leverages the KNN algorithm to construct hyperedges in hypergraphs, thereby addressing the limitations of spatial dependencies that traditional spatio-temporal convolutions are unable to model. 
% For the spatial shifts inherent in different node evolutions, the order of the hypergraph plays a pivotal role. 
As illustrated in Figure~\ref{fig04}, the effective order $k=3$ of the adaptive hypergraph significantly enhances model performance. On the BIKE-Outflow and PEMS03 datasets, STH-SepNet models based on BERT, GPT-2, LLAMA1B, and DeepSeek1.5B demonstrate that when $k=2$, the high-order structure of STH-SepNet degenerates into a pairwise relationship. The empirical results indicate that as the order 
$k$ increases, the model error initially decreases and then increases. This phenomenon is attributed to the fact that at $k=2$, pairwise relationships fail to capture the underlying dependencies, while higher-order hypergraph structures with increased 
$k$ order lead to overfitting of coupled interactions. That is, 
$k=3$ effectively characterizes evolving spatial dependencies.

\subsection{Computational Efficiency Analysis}
To analyze the advantages of decoupling in algorithmic efficiency, we test the STH-SepNet model with multiple large language models (LLMs) by comparing their GPU usage and training speed on different datasets using an NVIDIA A6000. Figure.\ref{fig05} shows that the STH-SepNet series outperforms TIMELLM in computational efficiency across most datasets. For instance, STH-SepNet (BERT) showed a GPU memory usage of just 24.6G and a training speed of 392 Epoch/s on the BIKE-Inflow/Outflow dataset, exhibiting superiority over both TIMELLM (BERT) and TIMELLM (GPT2). Furthermore, as the parameter size of LLMs increases, computational efficiency tends to decrease. However, larger model parameters do not enhance the accuracy of spatiotemporal predictions. This indicates that STH-SepNet (BERT) generally outperforms TIMELLM and larger STH-SepNet models in terms of GPU usage, training speed, and overall performance. This advantage stems from STH-SepNet’s decoupled processing of temporal features and its use of average pooling to extract global trend features (Eq.~\ref{avgpool}), which reduces GPU resource consumption and boosts efficiency.

% \begin{table}[htbp]
% % \small
% \setlength{\abovecaptionskip}{0pt}
% \setlength{\belowcaptionskip}{0pt} 
% \caption{Analysis of hyper-parameter $\gamma$ on hypergraph spatio-temporal module (LLMs:BERT).}
% \label{table03}
% \begin{tabular}{lcccc}
% \hline
% \hline
% \multirow{1}{*}{Model} & \multicolumn{2}{c}{BIKE-Outflow } & \multicolumn{2}{c}{PEMS03 } \\
% \cmidrule (lr){2-3} \cmidrule (lr){4-5}
% \multicolumn{1}{c}{} & MAE & RMSE & MAE & RMSE \\
% \midrule
% STH-SepNet-GNN ($\gamma=1$) & 5.47 & 13.36 & 21.11 & 34.52 \\
% STH-SepNet-MixGNN ($\gamma=0.5$) & \underline{5.39} & \underline{13.97} & \underline{21.09} & \underline{34.39} \\
% STH-SepNet ($\gamma=0$) & \textbf{ 5.33} & \textbf{14.23} & \textbf{21.03} & \textbf{34.17} \\
% \hline
% \hline
% \end{tabular}
% \end{table}

% \subsubsection{\textbf{Mixed-order spatio-temporal convolutional networks} }
% To investigate the influence of adaptive network structures within a hypergraph spatio-temporal module, we examine the STH-SepNet-MixGNN. This model integrates a mixed-order (2-order and 3-order) graph approach in its spatio-temporal module. As illustrated in Table \ref{table03}, STH-SepNet-MixGNN outperforms traditional adaptive graph methods (STH-SepNet-GNN) on BIKE-Outflow and PEMS03 datasets. This superior performance is attributed to the model's ability to capture complex, higher-order dependencies within spatio-temporal data, effectively combining these with real geographic networks.

\section{Conclusion}
This paper introduces STH-SepNet, a framework for spatio-temporal prediction that decouples temporal and spatial modeling through two specialized components: lightweight large language models for temporal dynamics and adaptive hypergraphs for spatial dependencies. 
By employing a spatio-temporal decoupling design, the ability of STH-SepNet to predict spatio-temporal data is significantly enhanced.
% By isolating these dimensions, the framework mitigates noise caused by joint modeling approaches.
Experimental results demonstrate improved accuracy across diverse datasets, including traffic networks
 (e.g., PEMS03, MAE: 21.03 vs. 26.84 for non-LLM variants) 
and urban mobility systems (e.g., BIKE-Outflow, MAE: 5.33 vs. 6.74 for LLM baselines). The adaptive hypergraph structure dynamically adjusts to spatial distribution shifts, such as policy-driven traffic pattern changes or sudden disruptions, enabling robust predictions in dynamic environments. 
The improved performance is attributed to the decoupled architecture, which allows temporal and spatial modules to focus on distinct patterns without mutual interference. Adaptive hypergraphs address the limitations of static graph structures by modeling higher-order interactions and real-time spatial drift, while lightweight LLMs efficiently capture temporal trends. This design is shown to generalize across datasets with varying scales and dynamics, as evidenced by consistent results in both small-scale (e.g., BIKE) and large-scale (e.g., METR-LA) scenarios. 

\textbf{Limitations and future work.}
While STH-SepNet demonstrates strong performance, its current design has limitations. The framework assumes temporal and spatial dependencies can be cleanly decoupled, which may not hold in scenarios where these dimensions are intrinsically intertwined (e.g., rapidly evolving events with coupled spatio-temporal causality). Additionally, the adaptive hypergraph’s reliance on real-time node feature updates could pose challenges in latency-critical applications, where computational overhead for dynamic hyperedge generation might limit responsiveness. In future work, we will tackle these constraints by delving into hybrid architectures that strike a balance between decoupling and controlled interaction mechanisms.

\section{Acknowledgments}
This work is supported by the National Key R{\&}D Program of China under Grant No. 2022ZD0120004, the Zhishan Youth Scholar Program, the National Natural Science Foundation of China under Grant Nos. 62233004, 62273090, and the Jiangsu Provincial Scientific Research Center of Applied Mathematics under Grant No. BK20233002.

\bibliographystyle{ACM-Reference-Format}
\bibliography{kdd_2019}

% 
% If your work has an appendix, this is the place to put it.

% \clearpage
\appendix

\section{Hypergraph Theory}
% \subsection{Supplement of theory and definition}
\renewcommand{\thetheorem}{A.\arabic{theorem}}

\begin{definition}  
\textbf{ (Hyperedge)} Given a high-order graph $H= (V,E)$, a hyperedge $e \in E$ is a non-empty subset of $V$. For each $e \in E$, $e \neq \emptyset$, and $e= (v_{i_1},v_{i_{2}},\dots, v_{i_k}),v_{i_j}\in V$~\cite{huang2021unignn}.
\end{definition}

\begin{definition}
\textbf{ (k-uniform Hyperedge)}
If a hyperedge $e \in E$ contains exactly $k$ vertices, then it is called a $k$-uniform hyperedge~\cite{huang2021unignn}. Formally, for each $e \in E$, $|e| = k$.
\end{definition}

\begin{definition}
\textbf{(k-hops neighborhoods)} Given a node $v_i$, its k-hops neighborhood $N_{k} (v_i)$ comprises all nodes that can be reached from $v_i$ via at most $k$ edges from $v_i$ .
\end{definition}

% \begin{theorem}[]
% For any $k \geq 2$, the $ (k-1)$-hops neighborhood of a node $v$, denoted as $N_{k-1} (v)$, corresponds to all nodes involved in the $k$-order hyperedges in $H_v^k$, if and only if the following conditions are satisfied: For each $w\in N_{k-1} (v)$, 
%  (1) Local Connectivity Condition: there exists at least one path from $v$ to $w$ consisting of at most $k-1$ hyperedges. 
%  (2) Hyperedge Coverage Condition: there exists a $k$-order hyperedge $e\in H_{v}^{k}$ such that $w\in e$ and $e$ contains $v, w$, and at most $k-2$ intermediate nodes. 
%  (3) Uniqueness Condition: if there exist multiple $k$-order hyperedges containing both $v$ and $w$, then these hyperedges must share the same set of intermediate nodes.
%  Formally:
% \begin{equation}
% w \in N_{k-1} (v) \iff \{v, F_1, F_2, \ldots, u_k, w\} \in H_v^k, 
% \end{equation}
% where $F_1, F_2, \ldots, u_k$ are intermediary nodes.
% \end{theorem}

\textbf{Proof for Theorem.~\ref{theorem01}}. The theorem will be proved by showing both directions of the equivalence.
\paragraph{ (1) Sufficiency ($\Rightarrow$):}
Assume $w\in N_{k-1} (v)$, we need to show that there exists a $k$-order hyperedge $e\in H_{v}^{k}$ such that $w\in e$ and $e$ contains $v, w$, and most $k-2$ intermediate nodes. By the local connectivity condition, there exists a path from $v$ to $w$ using at most $k-1$ hyperedges. Let this path be represented by the sequence of hyperedges $e_1, e_2, \dots, e_{k-1}$. Since each hyperedge can connect more than two nodes, we can construct a $k$-order hyperedge $e$ that includes $v$ and $w$ along with at most $k-2$ intermediate nodes. This satisfies the hyperedge coverage condition. If there are multiple such hyperedges, the uniqueness condition ensures that they share the same set of intermediate nodes, thus ensuring consistency.

\paragraph{ (2) Necessity ($\Leftarrow$):}
Assume there exists a k-order hyperedge $e\in H_{v}^{k}$ such that $w\in e$ and $e$ contains $v, w$, and at most $k-2$ intermediate nodes. We need to show that $w\in N_{k-1} (v)$. By definition, the hyperedge 
$e$ connects $v$ and $w$ through at most $k-2$ intermediate nodes. This implies that there is a path from $v$ to $w$ consisting of at most $k-1$ hyperedges (since the hyperedge itself can be considered as part of the path). Thus, $w$ is within the $ (k-1)$-hops neighborhood of $v$, satisfying the local connectivity condition. The hyperedge coverage condition is directly satisfied by the existence of $e$, and the uniqueness condition ensures that no other hyperedges contradict this structure.

We supplement the pseudocode for the hypergraph generation process in Algorithm.\ref{algorithm1} as follows.

\begin{algorithm}[h]
\setlength{\abovecaptionskip}{0pt}
\setlength{\belowcaptionskip}{3pt} 
\caption{Hyperedge Construction}
\label{algorithm1}
\begin{algorithmic}[1]
\Require Batch data [$B, L, N, F$], parameter $k$ of high-order 
\Ensure Constructed hyperedges and spatial interaction results

\State Initialize hyperedges set: $\text{hyperedges} = \emptyset$
\For{each node in $V(|V|=N)$}
    \State Find $k$ nearest neighbors using KNN
    \State Construct dynamic $k$-order hyperedge
    \State Add the hyperedge to $\text{hyperedges}$
\EndFor
\State Calculate spatial interaction based on Theorem 1
\State Construct adaptive hypergraph based on hyperedges‘ set in each batch data 

\end{algorithmic}
\end{algorithm}

\section{Experiment Settings and Results}
\renewcommand{\thetable}{B.\arabic{table}} 
\setcounter{table}{0}

\subsection{Datasets}
\begin{itemize}
\item BIKE-Inflow/Outflow: The dataset captures bicycle demand across 295 traffic nodes in New York, recorded hourly. The dataset spans from 2023-01-01 00:00 to 2024-01-01 23:00. 
\item PEMS03 dataset: The dataset contains traffic speed data from 358 stations in the California Highway System, with a 5-minute interval. The time range covers weekdays from 2008-01-01 00:00 to 2008-03-31 23:55:00. 
\item BJ500: The dataset consists of traffic speed information from 500 stations in the Beijing Highway System, also at 5-minute intervals. The dataset covers weekdays from 2020-07-01 00:00:00 2020-07-31 23:55:00. 
\item METR-LA: The dataset from the Los Angeles Metropolitan Transportation Authority contains average traffic
speed measured by 207 loop detectors on the highways of
Los Angeles County ranging from Mar 2012 to Jun 2012.
\end{itemize}

\subsection{Large Language Models}

\begin{table}[htbp]
\small
 \centering
\setlength{\abovecaptionskip}{0pt}
\setlength{\belowcaptionskip}{3pt} 
 \caption{Comparison of parameter sizes and dimensions across large language models}
 \label{tables1}
 \renewcommand{\arraystretch}{1} % 调整行距
 \setlength{\tabcolsep}{11pt} % 调整列间距
 \begin{tabular}{lcc}
 \hline \hline
 Model & Parameters & LLM Dimension \\
 \midrule
 BERT \footnotemark[1] & 110M & 768 \\
 GPT-2 \footnotemark[2] & 124M & 768 \\
 GPT-3 \footnotemark[3] & 7580M & 4096 \\
 LLAMA-1B \footnotemark[4] & 1230M & 2048 \\
 LLAMA-7B \footnotemark[5] & 6740M & 4096 \\
 LLAMA-8B \footnotemark[6] & 8000M & 4096 \\
 DeepSeek-Qwen1.5B \footnotemark[7] & 1500M & 1536 \\
 \hline \hline
 \end{tabular}
\end{table}
\footnotetext[1]{https://huggingface.co/google-bert/bert-base-uncased}
\footnotetext[2]{https://huggingface.co/openai-community/gpt2}
\footnotetext[3]{https://huggingface.co/TurkuNLP/gpt3-finnish-large}
\footnotetext[4]{https://huggingface.co/meta-llama/Llama-3.2-1B}
\footnotetext[5]{https://huggingface.co/huggyllama/llama-7b}
\footnotetext[6]{https://huggingface.co/meta-llama/Llama-3.1-8B-Instruct}
\footnotetext[7]{https://huggingface.co/deepseek-ai/DeepSeek-R1-Distill-Qwen-1.5B}

On the PEMS03 dataset, prefix prompts are designed as follows:

[Dataset Description] This data comes from the solar power plant power dataset PV, which consists of 69 nodes located near North Carolina. The values represent the normalized power output, with a time sampling granularity of 1 hour. The input data has been processed using mean pooling across nodes to characterize the overall features within the region. Please note that this dataset exhibits a clear periodic pattern, with significant power output from 8 AM to 5 PM, and zero power output at night.

[Task Instruction] Forecast the next $L$ steps given the previous $H$ steps.

[Statistical Information] The timestamp information is formatted as [month, day, hour, minutes]. The input time begins from $\langle \text{start time} \rangle$ to $\langle \text{end time} \rangle$, and the prediction time spans from \\ $\langle \text{start prediction time} \rangle$ to $\langle \text{end prediction time} \rangle$. The minimum value is $\langle \text{min value} \rangle$, the maximum value is $\langle \text{max value} \rangle$, and the median value is $\langle \text{median value} \rangle$. The trend of the input is either upward or downward. The top 5 lags are $\langle \text{lag values} \rangle$.

\section{Ablation Experiment Results}
\renewcommand{\thetable}{C.\arabic{table}} 
\setcounter{table}{0}
\renewcommand{\thefigure}{C.\arabic{figure}} 
\setcounter{figure}{0}

\subsection{Ablation Study (RQ3 and RQ4)}

\begin{table*}[htbp]
\small
\setlength{\abovecaptionskip}{0pt}
\setlength{\belowcaptionskip}{3pt} 
\caption{Performance comparison. Multivariate prediction results for different LLMs with a prediction horizon of 48 time steps and a fixed backtracking window of T=48. Bolded results indicate the best performance.}
\label{tables2}
 \renewcommand{\arraystretch}{1} % 调整行距
 \setlength{\tabcolsep}{9pt} % 调整列间距
\begin{tabular}{lccccllccll}
\hline
\hline
 Model & \multicolumn{2}{c}{BIKE-Inflow} & \multicolumn{2}{c}{BIKE-Outflow} & \multicolumn{2}{c}{PEMS03} & \multicolumn{2}{c}{BJ500} & \multicolumn{2}{c}{METR-LA}\\
 \cmidrule (lr){2-3} \cmidrule (lr){4-5} \cmidrule (lr){6-7}\cmidrule (lr){8-9} \cmidrule (lr){10-11}
 & MAE & RMSE & MAE & RMSE & MAE & RMSE & MAE & RMSE & MAE & RMSE \\
 \midrule
STH-SepNet-w/o (LLMs) & 5.47 & \textbf{14.01} & 5.83 & 15.17 & 26.84 & 43.44 & 6.24 & 10.81 & 10.31 & 18.26 \\
STH-SepNet (BERT)-w/i & \textbf{5.18} & 14.40 & 5.33 & 14.23 & \textbf{21.03} & \textbf{34.17} & \textbf{5.58} & 9.77 & \textbf{9.42} & \textbf{16.41} \\
STH-SepNet (GPT2)-w/i & 5.35 & 14.71 & 5.31 & 14.24 & 21.43 & 35.01 & 5.88 & 9.69 & \underline{9.57} & \underline{16.64} \\
STH-SepNet (GPT3)-w/i & 5.36 & 14.78 & 5.24 & 14.16 & \underline{21.13} & \underline{34.69} & 5.83 & 9.92 & 9.56 & 16.77 \\
STH-SepNet (LLAMA1B)-w/i & 5.36 & 14.80 & 5.29 & 14.20 & 21.37 & 34.92 & 6.03 & 10.08 & 9.64
 & 16.84 \\
STH-SepNet (LLAMA7B)-w/i & \textbf{5.18} & 14.24 & 5.34 & 14.31 & 21.52 & 35.27 & \underline{5.73} & \underline{9.58} & 9.72 & 16.90 \\
STH-SepNet (LLAMA8B)-w/i & 5.30 & 14.66 & \underline{5.28} & \underline{14.20} & 21.51 & 35.19 & 5.80 & \textbf{9.65} & 9.71 & 16.94 \\ 
STH-SepNet (DeepSeek-Qwen1.5B)-w/i & \underline{5.25} & \underline{14.48} & \textbf{5.27} & \textbf{14.19} & 21.39 & 34.96 & 5.91 & 9.94 & 9.65 & 16.89 \\
\hline
\hline
\end{tabular}
\end{table*}

We compare two model variants: one without large language models (LLMs) — referred to as STH-SepNet-w/o — which retains only the adaptive hypergraph and linear temporal layers, and another full framework — denoted as STH-SepNet (XXX)-w/i — equipped with various LLM backbones, including BERT, GPT, LLAMA, and Deepseek. As shown in Table~\ref{tables2}
On the PEMS03 dataset, which exhibits dynamic traffic patterns, the exclusion of LLMs results in significantly higher errors (MAE: 26.84, RMSE: 43.44) compared to the full model with a BERT backbone (MAE: 21.03, RMSE: 34.17), corresponding to increases of 21.7\% and 21.3\%, respectively. Similarly, on the BJ500 dataset, the BERT-equipped variant (MAE: 5.58) outperforms STH-SepNet-w/o (MAE: 6.24) , underscoring the LLM’s ability to enhance the performance of traffic data prediction. The results demonstrate the two key findings that LLMs facilitate spatio-temporal prediction and superior performance of our framework.

\subsection{Ablation Study for Gating Mechanism}
\begin{figure}[htbp]
 \centering
 \setlength{\abovecaptionskip}{0pt}
 \setlength{\belowcaptionskip}{-8pt} 
 \includegraphics[width=1\linewidth]{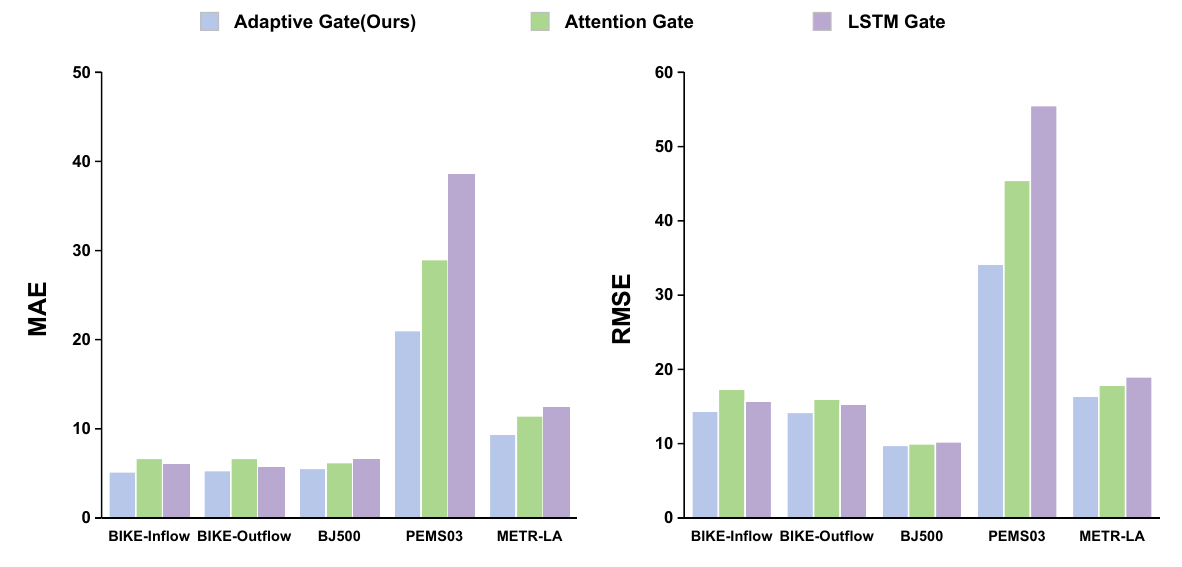}
\caption{Ablation studies: comparison of the three fusion mechanisms on various datasets. (LLMs:BERT).}
 \label{figs6}
\end{figure}

We systematically evaluate three fusion mechanisms across five datasets. Figure.~\ref{figs6} displays our proposed adaptive gating mechanism achieves superior prediction accuracy (MAE/RMSE) on all benchmarks. Notably, the adaptive gating exhibits particularly significant performance advantages in complex traffic scenarios like METR-LA and BJ500 (17.7\%–45.6\% MAE reduction), compared to cross-attention and LSTM-based gating mechanisms. Further analysis reveals distinct characteristics of alternative approaches: While cross-attention gating achieves suboptimal performance on PEMS03, its global attention computation introduces redundant feature interactions and demonstrates vulnerability to noise under sparse data conditions. The LSTM-based gating shows moderate temporal modeling capability in BIKE-Outflow prediction but fails to effectively capture the dynamic evolution of spatial features due to its unidirectional chain structure. 

The superiority of adaptive gating stems from its learnable weighting parameters that dynamically balance spatiotemporal feature contributions: 1) abrupt speed changes in METR-LA datasets, 2) tidal flow patterns in BIKE datasets. This elastic fusion strategy preserves the  advantages of decoupled design for heterogeneous features while parametrically compensating for potential information loss through dynamic weight adaptation.

\subsection{Effective Order of Adaptive Hypergraph}
Table \ref{tables5} illustrates that as the order of the adaptive hypergraph increases, the RMSE error of the model initially decreases and then increases. Specifically, on the BikeOutflow and PEMS03 datasets, when $k=3$, the STH-SepNet model achieves the smallest MSE values. This finding indicates that an appropriately chosen hypergraph order can significantly boost spatio-temporal prediction performance, thereby validating the effectiveness of the hypergraph structure in capturing spatial dependencies.

\begin{table}[htbp]
\small
\setlength{\abovecaptionskip}{3pt}
\setlength{\belowcaptionskip}{3pt} 
\caption{Analysis of effective order $k\in \{2,3,4,5\}$ on adaptive hypergraph. Performance metric (RMSE).}
\label{tables5}
 \renewcommand{\arraystretch}{1} % 调整行距
 \setlength{\tabcolsep}{6pt} % 调整列间距
\begin{tabular}{lccccc}
\hline \hline
\multicolumn{2}{l}{Dataset} & 2 & \multicolumn{1}{c}{3} & \multicolumn{1}{c}{4} & \multicolumn{1}{c}{5} \\ \hline
\multirow{4}{*}{BIKE-Outflow} & BERT & \multicolumn{1}{l}{\textbf{13.36}} & 14.23 & 14.89 & 15.37 \\
 & GPT2 & \multicolumn{1}{l}{14.48} & \textbf{14.24} & 15.29 & 15.58 \\
 & LLAMA1B & \multicolumn{1}{l}{14.80} & \textbf{14.20} & 15.54 & 16.19 \\
 & DeepSeek1.5B & 14.55 & \textbf{14.19} & 15.46 & 16.05 \\
 \hline
\multirow{4}{*}{PEMS03} & BERT & 34.52 & \textbf{34.17} & 35.81 & 37.84 \\
 & GPT2 & 35.78 & \textbf{35.01} & 36.22 & 37.74 \\
 & LLAMA1B & 35.87 & \textbf{34.92} & 37.16 & 39.07 \\
 & DeepSeek1.5B & 35.47 & \textbf{34.96} & 38.03 & 39.27 \\
 \hline \hline
\end{tabular}
\end{table}

\subsection{Ablation Study for Different Modules}
\begin{figure}[htbp]
 \centering
 \setlength{\abovecaptionskip}{0pt}
 \setlength{\belowcaptionskip}{-3pt} 
 \includegraphics[width=1\linewidth]{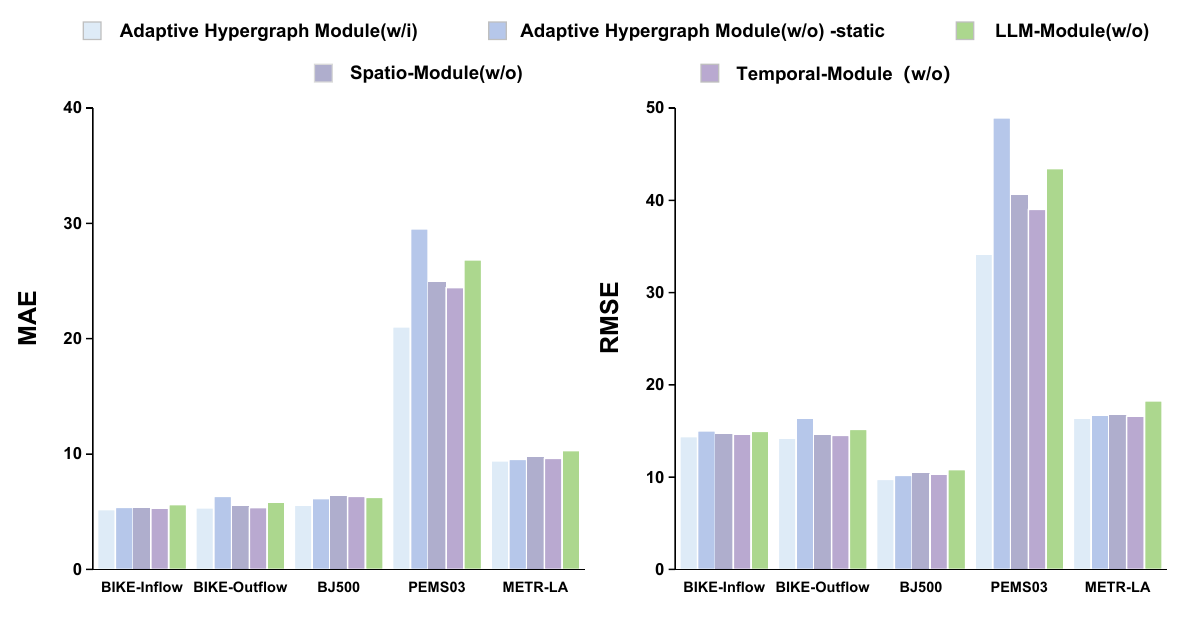}
\caption{Ablation studies: spatio, temporal, static graph,  LLMs, adaptive graph and hypergraph module. (LLMs:BERT).}
 \label{figs7}
\end{figure}

We conduct ablation studies on the SHT-SepNet model, which incorporates spatio-temporal modules, static graphs, LLMs, adaptive graphs, and hypergraph modules. Figure.~\ref{figs7} shows that the fully equipped model integrates the adaptive hypergraph module, the spatiotemporal module, and BERT as the language model module (LLM), demonstrates the most significant improvement in the two key performance metrics of MAE and RMSE. From a spatial perspective, the model with the adaptive hypergraph module (w/i) achieves remarkable performance on most datasets, though slightly inferior to the complete model. This suggests that the adaptive hypergraph module has a distinct advantage in capturing complex spatial relationships and can provide the model with crucial representational information. In contrast, when the adaptive hypergraph module is removed and only the static graph module is retained (w/o-static), there is a marked increase in MAE and RMSE for tasks such as BIKE-Inflow and BIKE-Outflow. For instance, in the BIKE-Inflow task, the MAE reaches 5.38, and the RMSE soars to 15.01. This indicates that the static graph struggles to adequately depict dynamic and complex spatial associations, thereby reducing the model's performance and reflecting the irreplaceable nature of the adaptive hypergraph module in learning dynamic spatial structures.

% We conduct ablation studies on the SHT-SepNet model, which includes spatio-temporal modules, static graphs, LLMs, adaptive graphs, and hypergraph modules. Figure.~\ref{figs7} shows the fully equipped model with the adaptive hypergraph module, spatiotemporal module, and BERT as the LLM achieves the best MAE and RMSE. The model with the adaptive hypergraph module (w/i) performs well on most datasets, indicating its advantage in capturing complex spatial relationships. However, when only the static graph module is used (w/o-static), MAE and RMSE increase significantly for tasks like BIKE-Inflow and BIKE-Outflow. For example, in the BIKE-Inflow task, MAE is 5.38 and RMSE is 15.01. This shows the static graph's limitations and highlights the adaptive hypergraph module's importance in dynamic spatial structure learning.

\subsection{Ablation Study for HGNNs}
To validate the adaptive hypergraph convolution method in spatio-temporal module, we compare HyperGCN~(Eq.~\ref{eq19} to \ref{eq21}), HyperGAT~\cite{huang2021unignn} and HyperSGAE~\cite{huang2021unignn} methods on BIKE-Inflow dataset. The experimental results are shown in Table \ref{tables3}. Under different methods, the adaptive hypergraph achieves better performance than the static graph in various indicators, which indicates that the adaptive hypergraph can more effectively capture and utilize the spatio-temporal relationship in the high-order convolutional network, thereby improving the accuracy and reliability of prediction.

\begin{table}[htbp]
\small
\setlength{\abovecaptionskip}{0pt}
\setlength{\belowcaptionskip}{3pt} 
\caption{Ablation studies: hypergraph neural networks like HyperGCN, HyperGAT and HyperSGAE.(LLMs: BERT).}
\label{tables3}
\renewcommand{\arraystretch}{1}  
\setlength{\tabcolsep}{4pt}  
\begin{tabular}{rcccccc}
\hline
\hline
& \multicolumn{2}{c}{HyperGCN} & \multicolumn{2}{c}{HyperGAT} & \multicolumn{2}{c}{HyperSGAE} \\
\cmidrule (lr){2-3} \cmidrule (lr){4-5} \cmidrule (lr){6-7}
& MAE        & RMSE        & MAE        & RMSE        & MAE         & RMSE        \\ 
\hline
Adaptive hypergraph & 5.18       & 14.40       & 5.01       & 13.97       & 5.01        & 13.93       \\
Static graph        & 5.38       & 15.01       & 5.47       & 14.79       & 5.42        & 14.65      \\
\hline
\hline
\end{tabular}
\end{table}

\end{document}